\documentclass[12pt,a4paper]{article}

\usepackage[utf8]{inputenc} 
\usepackage[english]{babel} 
\usepackage[T1]{fontenc} 
\usepackage{authblk} 
\usepackage{geometry} 
\usepackage{abstract} 
\usepackage{booktabs} 
\usepackage{graphicx} 
\usepackage{caption} 
\usepackage{multirow}
\usepackage{float} 
\usepackage{tabularx}           

\usepackage{parskip}        

\usepackage{hyperref}
\hypersetup{
    colorlinks=true,
    linkcolor=blue,
    filecolor=magenta,      
    urlcolor=cyan,
    citecolor=blue,
}

\title{\textbf{HealthQA-BR: A System-Wide Benchmark Reveals Critical Knowledge Gaps in Large Language Models}}
\author{Andrew Maranhão Ventura D'addario\textsuperscript{1}}
\affil{\small \textsuperscript{1}Independent Researcher}
\date{} 

\begin{document}

\maketitle

\begin{abstract}
\noindent
The evaluation of Large Language Models (LLMs) in healthcare has been dominated by physician-centric, English-language benchmarks, creating a dangerous illusion of competence that ignores the interprofessional nature of patient care. To provide a more holistic and realistic assessment, we introduce \textbf{HealthQA-BR}, the first large-scale, system-wide benchmark for Portuguese-speaking healthcare. Comprising 5,632 questions from Brazil’s national licensing and residency exams, it uniquely assesses knowledge not only in medicine and its specialties but also in nursing, dentistry, psychology, social work, and other allied health professions.

We conducted a rigorous zero-shot evaluation of over 20 leading LLMs. Our results reveal that while state-of-the-art models like GPT 4.1 achieve high overall accuracy (86.6\%), this top-line score masks alarming, previously unmeasured deficiencies. A granular analysis shows performance plummets from near-perfect in specialties like Ophthalmology (98.7\%) to barely passing in Neurosurgery (60.0\%) and, most notably, Social Work (68.4\%). This "spiky" knowledge profile is a systemic issue observed across all models, demonstrating that high-level scores are insufficient for safety validation. By publicly releasing HealthQA-BR and our evaluation suite, we provide a crucial tool to move beyond single-score evaluations and toward a more honest, granular audit of AI readiness for the entire healthcare team.
\end{abstract}

\section{Introduction}

The rapid ascent of Large Language Models (LLMs) has been marked by remarkable achievements on physician-centric benchmarks, with leading models now meeting or exceeding the passing threshold on high-stakes assessments like the United States Medical Licensing Examination (USMLE) \cite{usmle_review}. While impressive, these headline figures can create a dangerous illusion of competence. A single accuracy score, derived from a narrow set of tasks, risks masking critical, specialty-specific knowledge gaps and fundamentally misrepresents an AI's readiness for the complex reality of patient care.

This illusion of competence stems from two foundational biases in current evaluation paradigms. The first is a well-documented geographical and linguistic myopia. As recent work on the AfriMed-QA benchmark has shown, model performance can degrade significantly when tested outside of the high-resource, English-language contexts that dominate training data \cite{olatunji_afriqa}. Our work complements this geographical perspective by introducing a new, orthogonal dimension of evaluation: the professional diversity within a single, integrated healthcare system. Modern healthcare is not the domain of a single physician but a collaborative, interprofessional effort where nurses, pharmacists, therapists, and social workers contribute essential expertise. This system-wide reality of care—a team sport—has been largely ignored by physician-centric benchmarks, leaving a critical blind spot in our understanding of AI capabilities \cite{who_interprofessional}.

To address these gaps, we introduce \textbf{HealthQA-BR}, the first large-scale, system-wide benchmark for Portuguese-speaking healthcare. Comprising 5,632 questions from Brazil's national licensing and residency exams, HealthQA-BR moves beyond a monolithic focus on medicine to include dentistry, psychology, nursing, social work, and other allied health professions. Its design enables the granular, specialty-level analysis required to move past a single score and uncover the ``spiky'' knowledge profile of an LLM, pinpointing specific areas of both strength and alarming weakness.

Ultimately, this work is motivated by the ``last mile'' problem of AI adoption: the challenge of moving from passing an exam to earning the trust of clinicians, patients, and health systems \cite{psnet_safety, jrsm_trust}. True and safe adoption requires a rigorous, honest audit of a model's limitations. By providing a tool to identify and, consequently, remediate knowledge gaps across the entire healthcare team, we aim to help shift the paradigm from simply building models that can pass tests to cultivating AI that can function as a trustworthy and equitable partner in the collaborative endeavor of patient care.

\section{The HealthQA-BR Dataset}
\label{sec:dataset}

The foundation of this work is \textbf{HealthQA-BR}, a new, large-scale, and system-wide benchmark for evaluating the clinical knowledge and reasoning of Large Language Models (LLMs) in the context of Brazilian healthcare. The dataset is specifically designed to address the critical limitations of existing physician-centric benchmarks by encompassing the wide array of disciplines that constitute a modern, integrated healthcare system. It comprises \textbf{5,632} multiple-choice questions, meticulously sourced from high-stakes professional licensing and residency examinations in Brazil.

\begin{table}[htbp]
\centering
\small 
\caption{Composition of the HealthQA-BR Dataset. The dataset integrates questions from medical licensing, medical residency, and allied health residency examinations to create a comprehensive, system-wide benchmark.}
\label{tab:dataset_composition}
\begin{tabularx}{\textwidth}{l X r} 
\toprule
\textbf{Source} & \textbf{Description} & \textbf{No. of Questions} \\
\midrule
Revalida & Medical licensing exam for foreign-trained physicians & 1,777 \\
Enare - Médica & Medical residency entrance exam for physicians & 2,691 \\
Enare - Multiprofissional & Residency entrance exam for allied health professionals & 1,164 \\
\midrule
\textbf{Total} & & \textbf{5,632} \\
\bottomrule
\end{tabularx}
\end{table}

\subsection{Data Sources and Rationale}
\label{subsec:data_sources}

To ensure the clinical relevance, difficulty, and quality of our benchmark, we aggregated questions from two nationally recognized Brazilian examination bodies. The use of these sources guarantees that all questions have been developed and vetted by subject matter experts and reflect the core competencies required for professional practice in Brazil.

\begin{enumerate}
    \item \textbf{Revalida (Exame Nacional de Revalidação de Diplomas Médicos):} Administered by the Anísio Teixeira National Institute for Educational Studies and Research (INEP), the Revalida is a mandatory, high-stakes examination for physicians with foreign medical degrees who wish to practice medicine in Brazil \cite{revalida_source}. It is a notoriously challenging assessment designed to enforce a high standard of clinical competency, making it an ideal source for difficult benchmark questions. Our dataset includes \textbf{1,777 questions} from Revalida editions spanning 2011 to 2025.

    \item \textbf{Enare (Exame Nacional de Residência):} Organized by the Brazilian Hospital Services Company (Ebserh), Enare is a unified national examination for admission into a wide range of health residency programs across the country \cite{enare_source}. Critically, Ebserh manages federal university hospitals that are integral to Brazil's Unified Health System (SUS), the country's massive public healthcare provider. This direct link ensures that Enare's content is profoundly aligned with the real-world epidemiological profile, clinical challenges, and resource considerations of Brazil's public health landscape. Crucially, the Enare examination is offered in two distinct modalities for different professional groups, both of which are integral to our benchmark:
    \begin{itemize}
        \item \textbf{Enare - Residência Médica:} This examination stream is for medical doctors seeking specialization in areas such as Internal Medicine, General Surgery, Pediatrics, and Gynecology \& Obstetrics. It provides a deep dive into advanced medical knowledge and contributes \textbf{2,691 questions} to our benchmark.
        \item \textbf{Enare - Residência Multiprofissional:} This unique exam component establishes \mbox{HealthQA-BR} as a truly system-wide benchmark. It is a separate examination designed to assess candidates from allied health professions, including \textbf{Nursing, Psychology, Pharmacy, Physiotherapy, Social Work, and Nutrition}. These questions focus on the specific knowledge and interdisciplinary competencies required in a collaborative care setting, contributing \textbf{1,164 questions}.
    \end{itemize}
\end{enumerate}

The resulting composition of the HealthQA-BR dataset, combining all three sources, is summarized in Table \ref{tab:dataset_composition}. This multi-source, multi-disciplinary structure provides an unprecedented tool for evaluating an AI's readiness for the complexities of a complete healthcare ecosystem.

\subsection{Data Curation and Processing Pipeline}
\label{subsec:data_curation}

To transform raw examination documents into a high-fidelity benchmark, we designed and executed a meticulous, multi-stage data processing pipeline. This process was essential to ensure the quality, consistency, and analytical utility of the final dataset.

\begin{enumerate}
    \item \textbf{Parsing and Extraction:} Raw questions were extracted from publicly available PDF documents of past exams. This initial stage required robust automated scripts to handle significant variations in formatting, correct optical character recognition (OCR) artifacts, and resolve structural inconsistencies inherent in materials spanning over a decade of different layouts.

    \item \textbf{Curation and Integrity Validation:} Each extracted question, along with its options and correct answer, was subjected to a rigorous semi-automated and manual review. This critical step went beyond correcting simple parsing errors; we systematically cross-referenced all questions with the official examination answer keys and subsequent rectifications. This ensured that any questions that were officially nullified by the examination boards post-publication or had their answer keys amended were definitively removed from the benchmark, thereby preserving the dataset's clinical and academic integrity.

    \item \textbf{Advanced Deduplication:} To ensure the benchmark evaluates a broad range of knowledge rather than rote memorization, we implemented a sophisticated, two-tier deduplication strategy.
    \begin{itemize}
        \item First, all questions were canonicalized by removing whitespace and punctuation and then hashed to identify and eliminate exact duplicates.
        \item Second, to detect near-duplicates or paraphrased questions, we employed semantic similarity models. Questions flagged by the model as having high semantic overlap were then subjected to \textbf{manual expert review} to make a final determination. This hybrid approach ensures a uniquely clean dataset, free of both exact and substantive repetitions.
    \end{itemize}

    \item \textbf{Metadata Tagging and Granularity:} The most critical step for enabling fine-grained analysis was the assignment of rich metadata. Every entry in HealthQA-BR is tagged with its source (Revalida, Enare-Médica, Enare-Multiprofissional), examination year, and, most importantly, the relevant healthcare profession or medical subspecialty. We identified and applied over 30 distinct professional and subspecialty categories, including but not limited to General Medicine, Cardiology, Pediatric Nursing, Clinical Psychology, Dentistry (Endodontics), Pharmacy, and Physiotherapy. This comprehensive tagging is the key feature that allows for the precise identification of knowledge gaps across the healthcare system.

    \item \textbf{Final Structuring and Harmonization:} For analytical robustness and ease of use, we performed two final structuring steps.
    \begin{itemize}
        \item \textbf{Specialty Aggregation:} To ensure that all categories contained a sufficient number of questions for statistically meaningful analysis, we performed a strategic aggregation of closely related, low-sample subspecialties into broader, coherent groups (e.g., combining niche surgical fields into a single ``Surgical Subspecialties'' category).
        \item \textbf{Formatting:} A unique identifier was assigned to each question, and the final, fully-tagged dataset was serialized into the Apache Parquet format for computational efficiency and seamless integration with modern data analysis frameworks.
    \end{itemize}

    \item \textbf{Final Quality Assurance Audit:} To verify the end-to-end integrity of our pipeline, we conducted a final validation audit. A random, stratified sample of 82 questions (\textasciitilde1.5\% of the dataset), ensuring representation across different sources and specialties, was selected. Each sampled entry was manually inspected against its original source document to confirm the accuracy of the question text, options, correct answer, and all assigned metadata. This audit revealed no inconsistencies, confirming the high fidelity of the final HealthQA-BR dataset.
\end{enumerate}

\subsection{Dataset Characteristics}
\label{subsec:dataset_characteristics}

The final HealthQA-BR dataset consists of 5,632 unique multiple-choice questions, each with five answer choices and a single correct answer. The distribution of questions is 1,777 from Revalida and 3,855 from Enare (comprising 2,691 from medical residency and 1,164 from allied health residency). The dataset is structured in the Apache Parquet format for efficient processing, with clear fields for the question prompt, options, correct answer key, and all associated metadata.

The dataset's fine-grained metadata tagging allows for a detailed breakdown by professional area. The Enare - Residência Médica portion covers 37 distinct medical specialties, while the Enare - Residência Multiprofissional portion covers 15 allied health fields, as detailed in Table \ref{tab:medical_specialties} and Table \ref{tab:allied_health_specialties}.

\begin{table}[htbp]
\centering
\caption{Distribution of Questions by Medical Specialty from Enare - Residência Médica.}
\label{tab:medical_specialties}
\footnotesize 
\begin{tabularx}{\textwidth}{X r @{\hspace{2em}} X r}
\toprule
\textbf{Specialty} & \textbf{Count} & \textbf{Specialty} & \textbf{Count} \\
\midrule
Clínica Médica & 242 & Otorrinolaringologia & 66 \\
Cardiologia & 155 & Gastroenterologia & 64 \\
Medicina de Família, Preventiva e Saúde Coletiva & 146 & Foniatria & 58 \\
Cirurgia Geral & 141 & Cirurgia Vascular e Angiologia & 55 \\
Ginecologia e Obstetrícia & 111 & Endocrinologia e Metabologia & 51 \\
Radiologia e Diagnóstico por Imagem & 103 & Oncologia & 48 \\
Pediatria & 102 & Pneumologia & 42 \\
Hematologia e Hemoterapia & 100 & Medicina Intensiva & 42 \\
Homeopatia & 89 & Medicina do Adolescente & 41 \\
Neurologia & 89 & Medicina de Emergência & 40 \\
Cirurgia Plástica & 88 & Cirurgia do Aparelho Digestivo & 40 \\
Cirurgia Torácica & 83 & Alergia e Imunologia & 39 \\
Nefrologia & 82 & Medicina Física e Reabilitação & 27 \\
Patologia & 82 & Neurocirurgia & 25 \\
Psiquiatria & 81 & Cirurgia de Cabeça e Pescoço & 24 \\
Ortopedia e Traumatologia & 79 & Dermatologia & 14 \\
Oftalmologia & 78 & Anestesiologia & 10 \\
Infectologia & 75 & Reumatologia & 8 \\
Endoscopia Digestiva & 71 & & \\
\bottomrule
\end{tabularx}
\end{table}

\begin{table}[H]
\centering
\caption{Distribution of Questions by Allied Health Specialty from Enare - Residência Multiprofissional.}
\label{tab:allied_health_specialties}
\small
\begin{tabular}{lr@{\hspace{2em}}lr}
\toprule
\textbf{Specialty} & \textbf{Count} & \textbf{Specialty} & \textbf{Count} \\
\midrule
Saúde coletiva & 96 & Fisioterapia & 78 \\
Ciências biológicas & 81 & Psicologia & 75 \\
Odontologia & 80 & Farmácia & 74 \\
Biomedicina & 80 & Educação física & 73 \\
Fonoaudiologia & 79 & Física médica & 73 \\
Terapia ocupacional & 79 & Medicina veterinária & 70 \\
Serviço social & 79 & Nutrição & 69 \\
Enfermagem & 78 & & \\
\bottomrule
\end{tabular}
\end{table}

\section{Methodology}
\label{sec:methodology}

To rigorously assess the capabilities of Large Language Models (LLMs) on the \textbf{HealthQA-BR} benchmark, we designed a comprehensive and reproducible experimental protocol. Our methodology emphasizes a strict, controlled evaluation environment to ensure fair comparison across all models.

\subsection{Language Models Evaluated}
\label{subsec:models_evaluated}

We evaluated a diverse cohort of over 21 prominent LLMs, selected to represent the current state of the art. The selection includes leading proprietary models accessible via API, as well as a wide range of open-source models spanning various sizes and architectures. This allows for a thorough analysis of the impact of model scale, access, and training methodologies on performance in the healthcare domain. The full list of evaluated models, their developers, approximate parameter counts, and access methods are detailed in the supplementary data.

\subsection{Prompting and Inference}
\label{subsec:prompting}

All evaluations were conducted in a rigorous zero-shot setting to assess the models' inherent knowledge and reasoning capabilities without any task-specific training or examples. For each question in the \textbf{HealthQA-BR} dataset, the model was presented with a single, standardized prompt. The prompt contained the question text followed by the five multiple-choice options, labeled A through E. The model was instructed to return only the single letter corresponding to the correct answer.

To ensure consistency and comparability, inference parameters were held constant across all models. A complete list of all generation parameters is provided in the Supplementary Materials.

\subsection{Evaluation Metrics and Baselines}
\label{subsec:metrics}

The primary evaluation metric for this study is Accuracy, which represents the most charitable measure of a model's semantic knowledge on the benchmark. 

To contextualize the performance of the LLMs, we established two baselines:
\begin{itemize}
    \item \textbf{Most Common Answer:} For a five-option multiple-choice question, this baseline is 21.89\%.
    \item \textbf{Official Passing Score:} Where available, we use the official cut-off scores for the Revalida and Enare examinations (typically ranging from 60\% to 65\%) as a benchmark for minimum professional competence.
\end{itemize}

\section{Results}

This section presents the empirical results of evaluating over 21 Large Language Models on the HealthQA-BR benchmark. The analysis first provides a high-level comparison of all models across the three constituent examinations to establish a performance hierarchy. It then transitions to a fine-grained, specialty-level analysis of the top-performing model, GPT 4.1, to uncover the specific knowledge gaps that are the central focus of this study.

\subsection{Overall Model Performance Comparison}

\begin{figure}[H]
    \centering
    \includegraphics[width=1\linewidth]{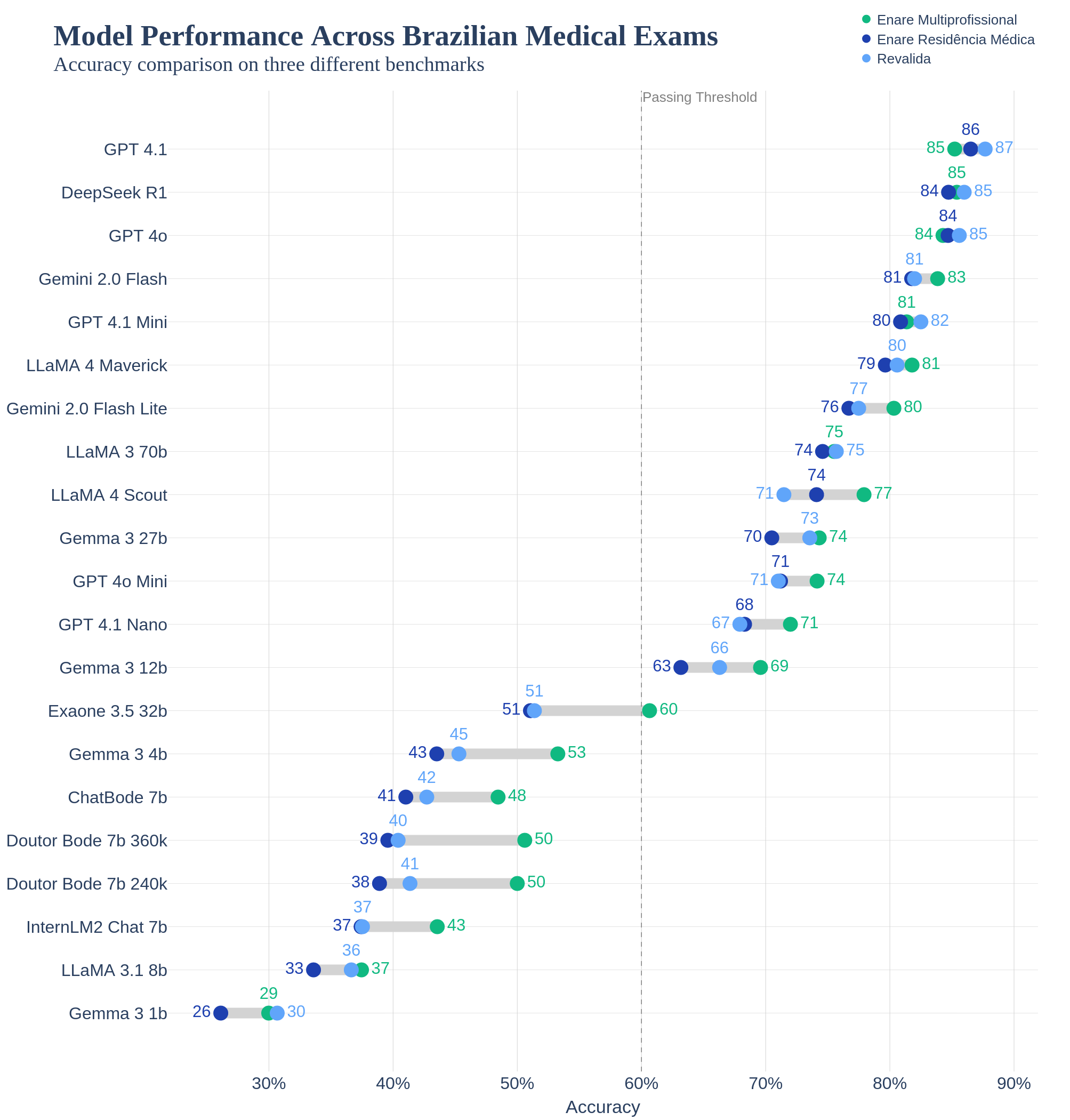}
    \caption{Model Performance Across Brazilian Medical Exams. The plot shows the accuracy of each model on the three sub-datasets (Revalida, Enare Residência Médica, Enare Multiprofissional) and the overall average. The dotted line indicates a typical passing threshold of 60\%.}
    \label{fig:general_results}
\end{figure}

The evaluation reveals a distinct and wide stratification of performance across the spectrum of LLMs, as illustrated in Figure \ref{fig:general_results} and detailed in Table \ref{tab:model_performance}. A clear hierarchy emerges, separating the models into distinct capability tiers.

\begin{table}[H]
	\centering 
	\caption{Model Performance Comparison Across Medical Exams}
	\label{tab:model_performance}
	\resizebox{\textwidth}{!}{%
    \Large 
    \renewcommand{\arraystretch}{1.2}%
	\begin{tabular}{llcccc}
	\toprule
	\textbf{Access Type} & \textbf{Model} & \textbf{Revalida} & \textbf{Enare Residência Médica} & \textbf{Enare Multiprofissional} & \textbf{Overall Accuracy} \\
	\midrule
	\multirow{7}{*}{\parbox{2cm}{Closed source}}
	& \textbf{GPT 4.1} & \textbf{0.8768} & \textbf{0.8651} & \textbf{0.8522} & \textbf{0.8661} \\
	& GPT 4.1 Mini & 0.8250 & 0.8086 & 0.8136 & 0.8148 \\
	& GPT 4.1 Nano & 0.6792 & 0.6830 & 0.7199 & 0.6895 \\
	& GPT 4o & 0.8559 & 0.8469 & 0.8428 & 0.8489 \\
	& GPT 4o Mini & 0.7102 & 0.7120 & 0.7414 & 0.7175 \\
	& Gemini 2.0 Flash & 0.8199 & 0.8175 & 0.8385 & 0.8226 \\
	& Gemini 2.0 Flash Lite & 0.7749 & 0.7670 & 0.8033 & 0.7770 \\
	\midrule
	\multirow{14}{*}{\parbox{3cm}{\raggedright Open source, open weights or permissible license}}
	& \textbf{DeepSeek R1} & \textbf{0.8599} & \textbf{0.8473} & \textbf{0.8540} & \textbf{0.8526} \\
	& LLaMA 4 Maverick & 0.8059 & 0.7964 & 0.8179 & 0.8038 \\
	& LLaMA 4 Scout & 0.7147 & 0.7410 & 0.7792 & 0.7406 \\
	& LLaMA 3.3 70b & 0.7569 & 0.7458 & 0.7552 & 0.7512 \\
	& LLaMA 3.1 8b & 0.3663 & 0.3359 & 0.3746 & 0.3535 \\
	& Exaone 3.5 32b & 0.5138 & 0.5106 & 0.6065 & 0.5314 \\
	& Gemma 3 27b & 0.7355 & 0.7049 & 0.7431 & 0.7225 \\
	& Gemma 3 12b & 0.6629 & 0.6317 & 0.6959 & 0.6548 \\
	& Gemma 3 4b & 0.4530 & 0.4352 & 0.5326 & 0.4609 \\
	& Gemma 3 1b & 0.3067 & 0.2612 & 0.2998 & 0.2836 \\
	& Doutor Bode 7b 360k & 0.4041 & 0.3958 & 0.5060 & 0.4212 \\
	& Doutor Bode 7b 240k & 0.4136 & 0.3891 & 0.5000 & 0.4197 \\
	& ChatBode 7b & 0.4271 & 0.4103 & 0.4845 & 0.4309 \\
	& InternLM2 Chat 7b & 0.3754 & 0.3742 & 0.4356 & 0.3873 \\
	\bottomrule
	\end{tabular}%
	}
\end{table}

\textbf{Top-Tier Performers:} At the apex are the state-of-the-art proprietary models. \textbf{GPT 4.1} establishes itself as the leading model, achieving the highest overall accuracy of \textbf{86.6\%}. It demonstrated consistently high performance across the board, scoring 87.7\% on Revalida, 86.5\% on Enare Residência Médica, and 85.2\% on Enare Multiprofissional. The premier open-source model, \textbf{DeepSeek R1}, proves to be highly competitive, achieving an overall accuracy of \textbf{85.3\%} and showing near-parity with the top closed-source model. Other models in this tier, including \textbf{GPT-4o (84.9\%)} and \textbf{Gemini 2.0 Flash (82.3\%)}, also comfortably surpass the professional passing threshold of approximately 60-65\% used in these examinations.

\textbf{Mid-Tier Performers:} A noticeable gap separates the top models from the next tier, which includes larger open-source models like \textbf{LLaMA 3 70b (75.1\%)} and \textbf{Gemma 3 27b (72.3\%)}. While these models perform credibly, they represent a clear step down in capability. Models such as \textbf{Gemma 3 12b (65.5\%)} perform at the margin, hovering just around the minimum competency benchmark.

\textbf{Lower-Tier Performers:} Performance drops precipitously for smaller and less advanced models. Models such as the 7-billion-parameter \textbf{Doutor Bode (42.1\%)}, \textbf{ChatBode (43.1\%)}, and \textbf{InternLM2 Chat (38.7\%)} fall significantly short of the passing grade. The smallest models evaluated, like \textbf{LLaMA 3.1 8b (35.4\%)} and \textbf{Gemma 3 1b (28.4\%)}, achieve scores that are not substantially better than random chance (21.9\%), indicating a fundamental lack of the specialized knowledge required for this domain.

\subsection{Granular Analysis of GPT 4.1 Reveals Specialty-Specific Knowledge Gaps}

While overall accuracy scores are informative, they can mask critical deficiencies. To test this paper's central hypothesis, we performed a granular analysis of the top-performing model, GPT 4.1, breaking down its accuracy by the specific medical and health specialties within the Enare exams.

It is crucial to emphasize that while the following analysis focuses on GPT 4.1 for clarity, \textbf{this pattern of uneven performance is a universal phenomenon observed across all evaluated models}. Even other state-of-the-art models like DeepSeek R1 and GPT-4o exhibit significant variance in their accuracy across different specialties. This indicates that knowledge gaps are not an idiosyncratic flaw of one model but a systemic issue in the current generation of LLMs.

\begin{figure}[H]
    \centering
    \includegraphics[width=1\linewidth]{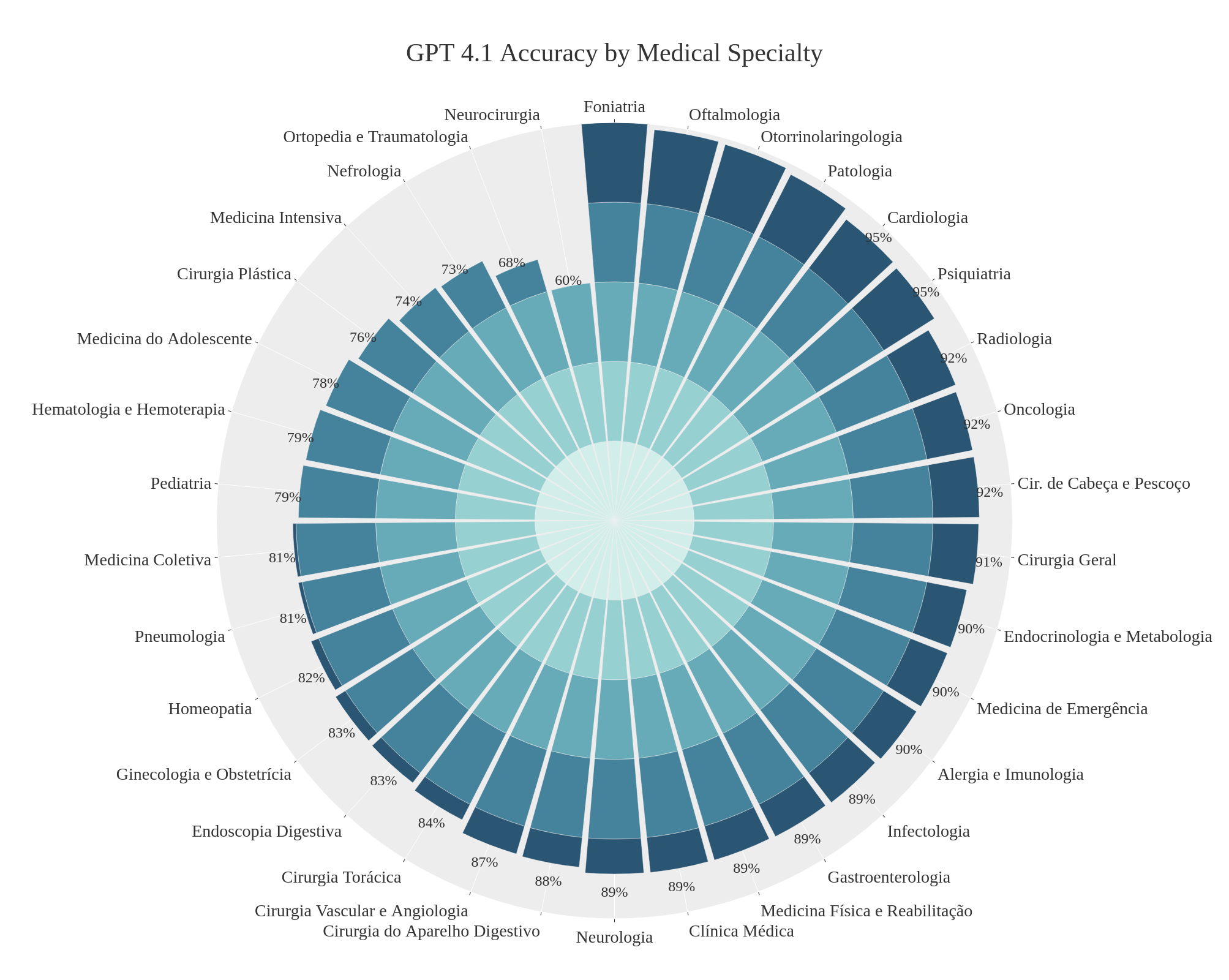}
    \caption{GPT 4.1 Accuracy by Medical Specialty. The chart displays the accuracy of the top-performing model on questions from the Enare - Residência Médica exam, broken down by 33 distinct medical specialties.}
    \label{fig:gpt_rm}
\end{figure}

As illustrated in Figure \ref{fig:gpt_rm}, GPT 4.1's performance on the Enare - Residência Médica exam reveals profound variability. The model demonstrates exceptional, near-perfect knowledge in several domains, achieving \textbf{100\% in Phoniatrics (Foniatria)}, \textbf{98.7\% in Ophthalmology}, and \textbf{97.6\% in Pathology}. In stark contrast, its accuracy plummets in other critical areas. Performance drops significantly in procedural and highly specialized fields, most notably to \textbf{68.4\% in Orthopedics and Traumatology} and a score of only \textbf{60.0\% in Neurosurgery}—barely at the passing threshold. Furthermore, it is noteworthy that foundational areas of primary and public health, such as \textbf{Pediatrics (79.4\%)} and \textbf{Collective Medicine (80.8\%)}, also scored significantly lower than the top-performing specialties.

\begin{figure}[H]
    \centering
    \includegraphics[width=1\linewidth]{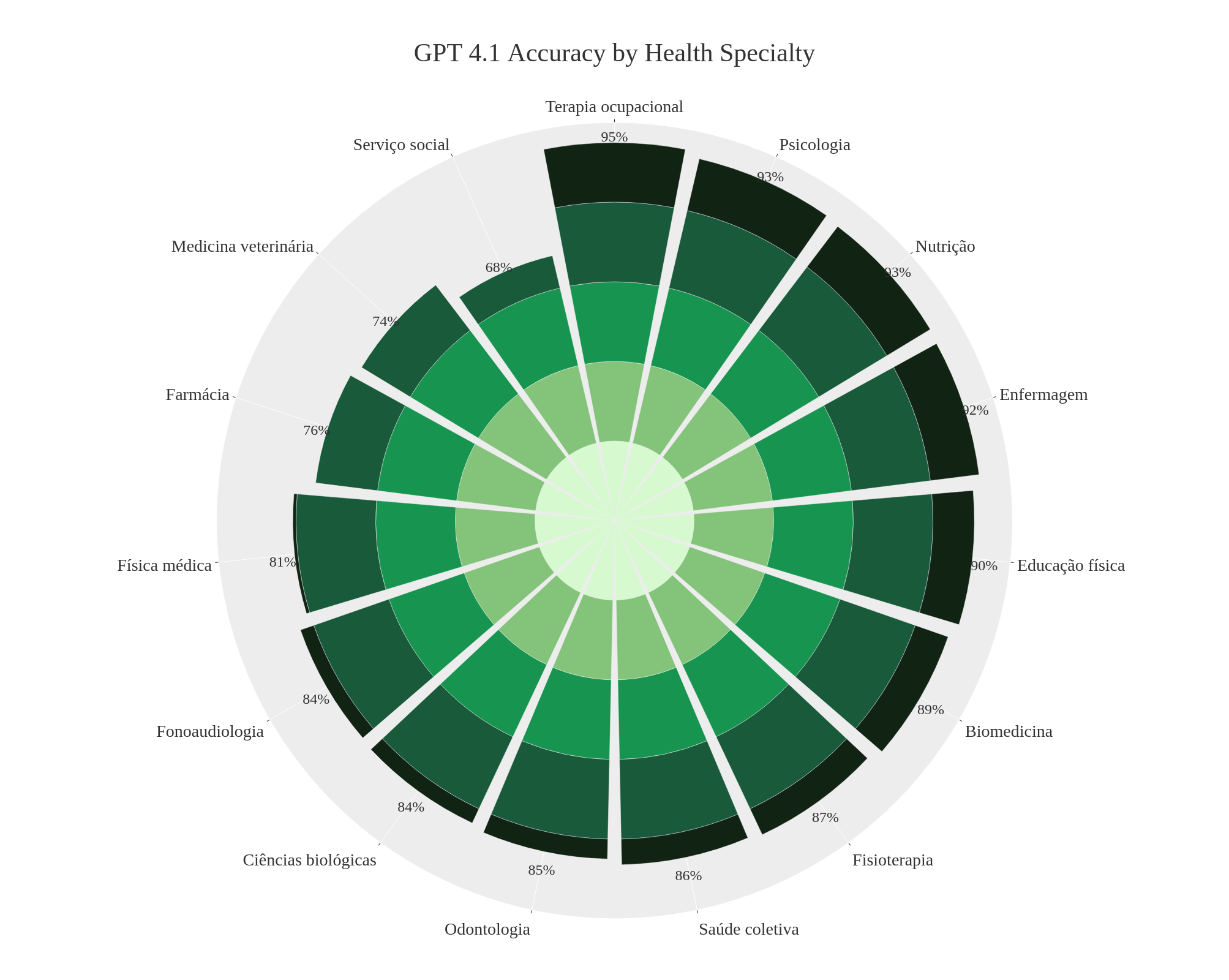}
    \caption{GPT 4.1 Accuracy by Health Specialty. The chart displays the accuracy of the top-performing model on questions from the Enare - Residência Multiprofissional exam, broken down by 15 allied health professions.}
    \label{fig:gpt_multi}
\end{figure}

This trend is reinforced by the analysis of the Enare - Residência Multiprofissional dataset, a unique contribution of this work, as shown in Figure \ref{fig:gpt_multi}. GPT 4.1 performs with very high accuracy in \textbf{Occupational Therapy (94.9\%)}, \textbf{Psychology (93.3\%)}, and \textbf{Nursing (92.3\%)}. However, the model exhibits its most significant knowledge gap in \textbf{Social Work}, with an accuracy of only \textbf{68.4\%}. Performance is also notably lower in \textbf{Veterinary Medicine (74.3\%)} and \textbf{Pharmacy (75.7\%)} compared to other allied health fields. The particularly low score in Social Work, a discipline integral to public health systems like Brazil's Unified Health System (SUS), is especially concerning and highlights a critical capability gap with direct relevance to public health delivery.

\section{Discussion}

Healthcare is fundamentally a team sport. The successful diagnosis, treatment, and holistic care of a patient does not rest on a single physician but on the coordinated expertise of a diverse team of professionals, including nurses, pharmacists, therapists, and social workers. For an Artificial Intelligence to become a valuable and safe member of this team, it cannot simply be a star player in a few select positions. However, the prevailing approach to evaluating medical AI has often been physician-centric, akin to scouting a single player while ignoring the needs and dynamics of the entire team. This study, through the system-wide lens of HealthQA-BR, provides the first comprehensive "team tryout" for modern LLMs. The results are clear: while today's top models possess star-level talent in certain domains, they are not yet versatile enough to play effectively and safely across the entire field.

The discovery of this "spiky" and uneven knowledge profile is the central finding of our study. It is crucial to note that this is not an isolated phenomenon. Our results are strongly corroborated by findings from the AfriMed-QA dataset, which also reported significant performance variability across different medical specialties and, notably, across different African countries (1). The combined evidence from both benchmarks suggests that uneven knowledge is a systemic issue in the current generation of LLMs, making granular, sub-domain analysis an absolute necessity for safe validation.

The particularly low scores in disciplines integral to public health, such as Social Work (68.4\%) and Collective Medicine (80.8\%), are especially concerning. This may point to a systemic bias in training data that over-represents specialized, high-income medicine at the expense of community-focused and preventative care. This aligns with concerns raised in the AfriMed-QA study regarding the readiness of LLMs for low-resource settings, where public health infrastructure is paramount. Furthermore, our finding that smaller, open-source models perform poorly mirrors their conclusions, highlighting a significant barrier to the equitable deployment of AI in regions that cannot rely on large, proprietary models (1).

The granular analysis presented in this work should not be viewed merely as a critique but as a constructive, strategic playbook for improvement. Moving beyond a single "win-loss" record—the overall accuracy score—our specialty-level breakdown serves as a detailed scouting report. This report enables two crucial pathways for progress. First, it allows for \textbf{focused initiatives}, empowering developers to see precisely where a model is weak (e.g., Orthopedics, Social Work) and requires targeted "coaching," whether through specialized fine-tuning or the integration of curated knowledge via Retrieval-Augmented Generation (RAG). Second, the universal nature of these deficiencies, particularly the underperformance in public health-related fields, suggests \textbf{pathways for general improvement}. It points to a systemic bias in current LLM training paradigms, which likely over-represent specialized medicine at the expense of primary care and community-focused disciplines. Addressing this requires building more balanced, multi-professional training corpora that reflect the full, integrated reality of a healthcare system.

Ultimately, the path from a high benchmark score to safe clinical deployment is not direct. It is contingent on actively remediating the knowledge gaps identified in this work. This represents the imperative of remediation for true and safe adoption. To deploy a model with a 68\% accuracy in Social Work or 60\% in Neurosurgery under the banner of an "87\% accurate medical AI" would be profoundly irresponsible. Such an action would not only risk suboptimal or even harmful outcomes for patients whose needs fall into one of the model's blind spots, but it would also deservedly erode the trust of providers. Therefore, the true and safe adoption of AI across healthcare systems requires an unwavering commitment to identifying and closing these knowledge gaps. This ensures that AI technology serves all patients and providers equitably, strengthening the entire healthcare team rather than just a few of its most prominent members.

We acknowledge that the present study has limitations that mirror broader trends in the field of medical AI evaluation. As highlighted in the systematic review by Bedi et al. \cite{bedi_jama}, our reliance on a multiple-choice question (MCQ) format to assess performance and our primary use of accuracy as the core evaluation metric represent a narrow approach. However, we posit that this focused methodology is an indispensable foundational step. The assessment of core clinical knowledge is a fundamental precursor to any subsequent, more complex evaluation of an LLM's utility in real-world clinical workflows. A model cannot be expected to perform sophisticated tasks like diagnostic reasoning or patient communication if it lacks the basic knowledge of the domain. Furthermore, the MCQ format, by its very structure, inherently limits the range of viable evaluation metrics; for a task with a single, verifiably correct answer, accuracy remains the most direct and unambiguous measure of semantic knowledge. While future work must incorporate broader metrics to assess fairness, robustness, and deployment considerations, establishing this validated baseline of knowledge is a crucial prerequisite for any such investigation.

\section{Conclusion}

The evaluation of Large Language Models in medicine has long been hampered by a narrow, often physician-centric focus that fails to capture the complexity of modern healthcare delivery. In response to this critical limitation, this paper introduced and deployed \textbf{HealthQA-BR}, the first large-scale, system-wide benchmark for Brazilian healthcare. By sourcing thousands of questions from Brazil's high-stakes licensing and residency exams across medicine, nursing, psychology, and other allied health professions, this work enabled a rigorous, multi-faceted evaluation of over 21 leading LLMs, moving beyond a single monolithic score to a more revealing, granular analysis.

Our central finding is that while state-of-the-art models can achieve overall accuracies that meet or exceed professional standards, this high-level performance represents a dangerous illusion of competence. The most critical discovery of this work is the "spiky" and uneven knowledge profile hidden beneath these scores—a profile characterized by excellence in some domains, such as Ophthalmology and Psychology, but alarming deficiencies in others, like Neurosurgery and Social Work. These are not idiosyncratic flaws but systemic issues observed across models, proving that granular, specialty-level validation is not an optional extra but a fundamental prerequisite for assessing the safety and reliability of any AI intended for clinical use.

The evidence presented herein demands a paradigm shift in the validation of medical AI. The field must evolve beyond the simplistic pursuit of ever-higher scores on narrow benchmarks and embrace a more sophisticated and responsible approach. We must transition from asking "How accurate is the model?" to asking the crucial subsequent questions: "Where is the model accurate? Where does it fail? What are its knowledge biases? And is it safe for the entire healthcare team?" This requires a commitment to embracing complexity, prioritizing transparency, and valuing a deep understanding of a model's limitations over the allure of a single, impressive number.

To catalyze this global effort, we are publicly releasing the \textbf{HealthQA-BR} benchmark, as well as our complete evaluation suite. We offer these resources to empower the international research community to conduct similar system-wide audits and to help determine if the knowledge gaps we have identified are universal. The ultimate goal must be to move beyond building models that can simply pass exams and toward cultivating AI that can function as a truly trustworthy, equitable, and reliable partner in the complex, collaborative endeavor of patient care. The future of medical AI must be built on a foundation of comprehensive knowledge, radical transparency, and an unwavering commitment to the safety and care of both patients and providers.

\section{Acknowledgments}
This work was supported by the Brazilian Ministry of Health (MoH/DECIT) in partnership with the National Council for Scientific and Technological Development (CNPq) [grant number 400757/2024-9] and the Gates Foundation. The Author Accepted Manuscript version arising from this submission will be published under a Creative Commons Attribution 4.0 Generic License.


\end{document}